\documentclass[letterpaper, 10 pt, conference]{ieeeconf}  

\IEEEoverridecommandlockouts                              

\usepackage{graphicx}
\usepackage{amsmath}
\usepackage{amssymb}
\usepackage{booktabs}
\usepackage{multirow}
\let\labelindent\relax
\usepackage{enumitem}
\usepackage{bbding}
\usepackage{cancel}
\usepackage[pagebackref,breaklinks,colorlinks]{hyperref}

\def\eg{\textit{e.g.}}
\def\ie{\textit{i.e.}}

\newcommand{\model}{SUIT}

\renewcommand{\paragraph}[1]{\vspace{-1.5mm}{\flushleft\textbf{#1}}} 

\title{\LARGE \bf
SUIT: Learning Significance-guided Information \\ for 3D Temporal Detection
}

\author{Zheyuan Zhou$^{1}$, Jiachen Lu$^{1}$, Yihan Zeng$^{2}$, Hang Xu$^{2}$, Li Zhang$^{1 \dag}$
\thanks{$^{1}$Zheyuan Zhou ({\tt \small zheyuanzhou20@fudan.edu.cn}), Jiachen Lu and Li Zhang ({\tt \small lizhangfd@fudan.edu.cn}) are with Fudan University.
}
\thanks{$^{2}$Yihan Zeng and Hang Xu are with Huawei Noah's Ark Lab.}
\thanks{$^{\dag}$Li Zhang is the corresponding author with School of Data Science, Fudan University.}
}

\begin{document}

\maketitle
\thispagestyle{empty}
\pagestyle{empty}

\begin{abstract}
3D object detection from LiDAR point cloud is of critical importance for autonomous driving and robotics. 
While sequential point cloud has the potential to enhance 3D perception through temporal information, utilizing these temporal features effectively and efficiently  remains a challenging problem. 
Based on the observation that the foreground information is sparsely distributed in LiDAR scenes, we believe sufficient knowledge can be provided by sparse format rather than dense maps. 
To this end, we propose to learn Significance-gUided Information for 3D Temporal detection (SUIT), which simplifies temporal information as sparse features for information fusion across frames. 
Specifically, we first introduce a significant sampling mechanism that extracts information-rich yet sparse features based on predicted object centroids. 
On top of that, we present an explicit geometric transformation learning technique, which learns the object-centric transformations among sparse features across frames. 
We evaluate our method on large-scale nuScenes and Waymo dataset, where our SUIT not only significantly reduces the memory and computation cost of temporal fusion, but also performs well over the state-of-the-art baselines.

%
\end{abstract}

\section{Introduction}
\label{sec:intro}

3D object detection with LiDAR point cloud is a fundamental task in autonomous driving~\cite{zhou2018voxelnet, yan2018second, yang2019std, shi2019pointrcnn, lang2019pointpillars, yang20203dssd,yin2020center,shi2020pv,deng2021voxel,fan2021rangedet,hu2022afdetv2}.
Since the LiDAR sensor collects data continuously, the sequential point cloud frames can provide additional information compared to a sparse single frame, which encourages the development of multi-frame detectors~\cite{20213D, luo2022transpillars, xu2022int,zhou2022centerformer}.
However, the introduction of temporal clues inevitably increases the memory burden and computation cost when storing and fusion the information cross frames.
Besides, the temporal information of sequential data is unaligned due to the rapid movement of objects in scenes cross time, which results in invalid fusion.
To tackle those issues, it requires a rethink of the temporal information learning for an effective yet efficient multi-frame detector.

\begin{figure}[ht]
	\begin{center}
		\includegraphics[width=1.0\linewidth]{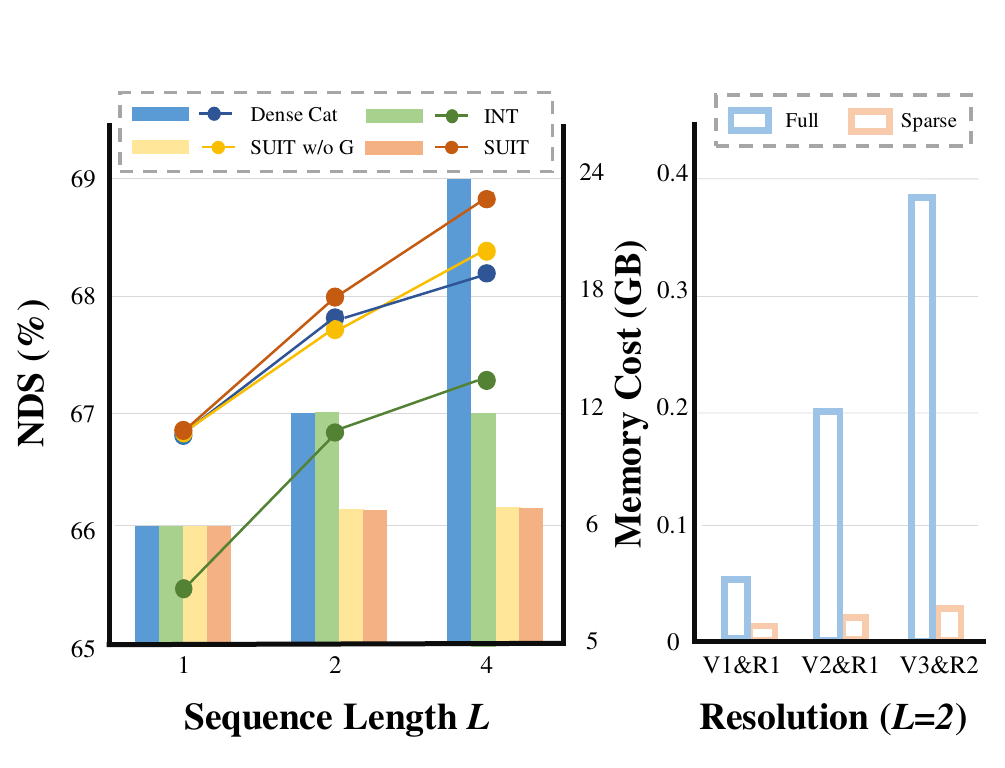}
	\end{center}
	\caption{Comparison of performance and memory cost. Our SUIT not only reduces the memory cost of temporal fusion with sparse features over different sequence lengths and resolutions but also achieves better temporal feature alignment with geometry transformation (G). 
	$\text{V1}$, $\text{V2}$, $\text{V3}$ represent voxel size of $[0.2m, 0.2m]$, $[0.1m, 0.1m]$, $[0.075m, 0.075m]$.
	$\text{R1}$, $\text{R2}$ represent point cloud range of $[-51.2m, 51.2m]$ and $[-54.0m, 54.0m]$.
	}
	\label{fig-intro}
\end{figure}


Previous works have shown success in improving multi-frame fusion for 3D object detection. Direct point-level fusion methods, such as concatenating multi-frame point clouds, have been explored with positive results \cite{caesar2020nuscenes, 2021Auto4D, yan2018second}. Meanwhile, feature-level fusion methods, which leverage the Bird's-eye-view feature maps of multiple frames, have also achieved considerable improvements \cite{2020An,yin2020lidar, yuan2020tctr, 20213D, luo2022transpillars, xu2022int}. Recently, those feature-level fusion methods have focused on learning temporal correlation between dense features using LSTM \cite{huang2020lstm}, GRU \cite{yin2020lidar}, or Transformer \cite{yuan2020tctr}. 



Despite the performance boost, current methods can not escape from utilizing the whole BEV feature maps at different times, resulting in high memory cost and computation waste by the repeated information across frames.
Previous work ~\cite{sun2021rsn} has proved that using range images only can achieve remarkable performance.
This observation inspired us that the whole dense BEV feature map is redundant as historical information since only the foreground areas swept by the LiDAR have clues that can be effectively resolved and exploited by the detector.
Furthermore, the point clouds are subject to discontinuities and inconsistencies during the acquisition process, leaving a huge gap for improvement between the alignment of multi-frame features. 
Though ~\cite{xu2022int} have proposed a way to mitigate the long-sequence memory usage, it still requires a dense feature from the past times.
Besides, it adopts simple concatenation to fuse temporal fusion without consideration of the explicit exploration of the unalignment of information between frames. 
By contrast, our method only exploits the slight sparse feature rather than complete BEV feature maps.
Therefore, memory consumption can be drastically whittled down whilst preserving most of the object information.
Based on these significant features with object-dense information, we can explicitly extract their object-centric correlation.

Under these observations, we propose a multi-frame 3D detector: {\bf\em Significance-guided 3D Temporal Detector (\model)}.
\model\ introduces two components: 1) {\em Significant sampling} and 2) {\em Explicit geometric transformation learning} to better answer two questions raised above: I) \textbf{Which} features to attend and II) \textbf{How} to aggregate.
Based on objects predicted by past frames, a significant sampling technique only saves features of objects information-rich in the past.
It promotes the medium between multiple frames which enables only a modicum of features will be maintained as the multi-frame medium in memory but still reserving enough information from past frames.
On the other hand, the explicit geometric transformation learning technique learns the object-centric transformation of these sparse features among frames with explicit object transformation supervision.
The joint efforts of both components drastically whittle away the computation burden under long-time dependence but also facilitate the learning of geometric transformation.

To prove the feasibility of \model, we conduct extensive multi-frame experiments on a large-scale automatic driving dataset nuScenes~\cite{caesar2020nuscenes} and Waymo~\cite{sun2020scalability}.
Figure~\ref{fig-intro} shows the accuracy v.s. time sequence length and the cost v.s. times sequence length.
Our \model\ achieves a larger boost as the time sequence extends whilst keeping a lower memory consumption increment.

In brief, our contributions can be summarized as follows: 
\begin{itemize}[noitemsep,nolistsep]
    \item We propose a multi-frame detector framework \model\ based on Significant sampling and Explicit geometric transformation learning.
    \item The proposed \model\ better answers the two basis: adequate medium to preserve cross-frame information and better geometric transformation understanding.
    \item Significant sampling sets significant features as medium-saving cross-frame information manifests its advantage of memory consumption and makes it possible for object-centric geometric transformation learning.
    Explicit geometric transformation learning modules realize the second basis thus promoting multi-frame estimation accuracy.
    \item Extensive experiments on nuScenes~\cite{caesar2020nuscenes} and Waymo~\cite{sun2020scalability} prove the effectiveness and efficiency of our \model.
\end{itemize}

\section{Related Work}
\label{sec:related}

\noindent \textbf{Single-Frame 3D Object Detection.}
Existing works on 3D object detection with point clouds alone can be mainly divided into two categories: point-based and grid-based.
Point-based methods~\cite{shi2019pointrcnn, yang2019std, yang20203dssd, shi2020point, zhou2020joint, zheng2021se, yu2022rotationally} directly learn the 3D representation of the scene by applying PointNet~\cite{qi2017pointnet, qi2017pointnet++} to process the irregular point cloud and generate the final proposal for each point.
These methods can preserve accurate geometry information from the raw points.
However, due to a large amount of time spent on organizing the irregular points~\
\cite{liu2019point}, operation on large-scale points
will lead to non-travail costs on computation and memory.

The grid-based schemes~\cite{zhou2018voxelnet, yan2018second, zhu2019class, lang2019pointpillars, hu2020you, yin2020center, hu2022afdetv2} first convert sparse point clouds into grid forms using 3D voxels or 2D pillars and then utilize the 3D/2D convolution networks to process the fixed size input.
VoxelNet~\cite{zhou2018voxelnet} is the first end-to-end voxel-based 3D object detector, which uses PointNet~\cite{qi2017pointnet, qi2017pointnet++} to encode the points within a voxel.
SECOND~\cite{yan2018second} further introduces 3D sparse convolution to tackle a large number of empty voxels in the outdoor scene for acceleration.
PointPillars~\cite{lang2019pointpillars} uses pillars rather than voxels to divide the point cloud.
Compared to point-based methods, grid-based methods are more efficient in computation and easier to deploy on autonomous driving vehicles.
In this work, we choose the popular grid-based CenterPoint~\cite{yin2020center} as our baseline model.

\begin{figure*}[t]
	\begin{center}
		\includegraphics[width=1.0\linewidth]{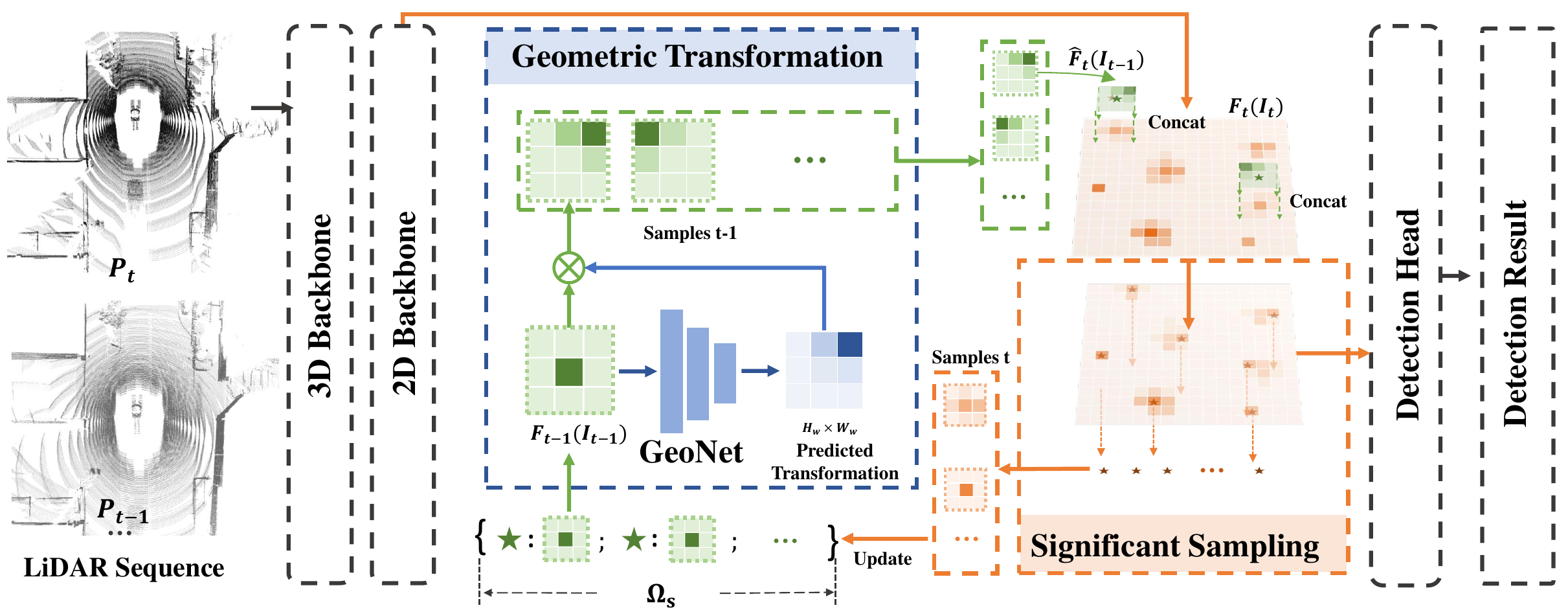}
	\end{center}
	\caption{Architecture overview. 
	Our multi-frame pipeline is positioned between the 2D backbone and detection head, which   comprises two key components: geometric transformation and significant sampling. Specifically, 
        geometric transformation aligns features across different time frames, while significant sampling identifies potential targets in the feature map and reduces the impact of irrelevant objects on the detection results.
	}
	\label{fig-arch}
\end{figure*}

\noindent \textbf{Multi-Frame 3D Object Detection.}
While handling point cloud sequence, a straightforward method is to concatenate points from the sweep~\cite{caesar2020nuscenes, 2021Auto4D, yan2018second}, which densifies the point cloud but ignores the temporal correlation between the neighbor sweeps. 
Although it's simple, the rapid growth of computation and memory usage makes it unable to handle long time series.
Instead, some recent approaches~\cite{2020An, yuan2020tctr, 20213D, luo2022transpillars, xu2022int, chen2022mppnet}, try to model the temporal interaction at the feature level.
The point cloud from the different frames will be voxelization and encoded independently and merged at the Bird's-eve-view features.
However, the point cloud data is read and processed repeatedly, and the sequence length is still bounded by its memory burden. 
3DVID~\cite{yin2020lidar} introduces a message-passing network and a GRU-based solution to aggregate the spatiotemporal information.
TCTR~\cite{yuan2020tctr} uses a transformer to fuse the middle-level feature, which further explores the spatial, temporal, and channel correlations among different frames.
3D-MAN~\cite{20213D} stores a series of proposals and their corresponding feature from the historical frame in a memory bank. 
A transformer is adopted to align and encode these features from different perspectives together with the current frame.
MPPNet~\cite{chen2022mppnet} is a two-stage point-based method, which uses a series of proxy points for multi-frame feature encoding and integrating. 
A series of self and global attention propagate the previous features in the whole trajectory.
INT~\cite{xu2022int} extends the storage of a fixed number of frames into an infinite by an on-stream training and prediction framework.
Compared to early works, our method explores the necessity of dense features, as well as the interaction between the neighbor frames.

\section{Preliminaries}
\paragraph{Single-Frame 3D Detector. }
A common voxel-based 3D object detection pipeline can be summarized as voxelization, 3D backbone, 2D backbone, and 3D detection heads.
Point cloud will be voxelized after voxelization and be represented as Bird's-eye-view features $F$ after the 3D\&2D backbone.
Our single-frame 3D detector baseline is implemented by the first stage of a commonly used 3D object detection head \cite{yin2020center}.
Before heads, feature map $F_t(I_t)\in\mathbb{R}^{H\times W}$ are produced by 2D backbone with input point cloud $I_t$.
For the popular adopted center head, the center locations of any probable objects are shown at the peak of a heatmap, as well as the categories of objects. 
Fundamentally, we use Bird's-eye-view coordinate $\Omega=\{r|r=(x,y)\}$ to represent the location where $x,y$ are restricted on the BEV feature map $F$.
Then, we define the random variable $R_t\in \mathbb{R}^{H\times W}$ to represent the probability of an object being at a given location at current time $t$.
By knowledge, center heatmap heads produce the following heatmap
\begin{equation}
\label{equ:single_det}
    p(R_{t}|I_{t}).
\end{equation}
\paragraph{Multi-Frame 3D Detector. }
The difference between single-frame and feature-level multi-frame detectors happens after 2D Backbone.
Our multi-frame 3D detector shares the same architecture single-frame 3D detector including voxelization, 3D backbone, and detection head.
But, our feature-level multi-frame 3D detector aggregates a temporal unified feature from the past features $F_{t}(I_t), F_{t-1}(I_{t-1}),\cdots$ after 2D backbone and before detection head.
Considering the nonalignment among these past features, the aggregated temporal unified feature can be modeled as follows.

Our target is to learn the temporal unified feature $\hat{F}_t(I_t, I_{t-1},\cdots)$.
The hat notation means the feature map is predicted by temporal transformation.
In convenience, we first introduce a 2-frame aggregation $f_2$, \ie{}
\begin{equation}
    \hat{F}_t(I_t,I_{t-1})=f_{2}\left[F_t(I_t), F_{t-1}(I_{t-1})\right].
\end{equation}

Noted that, to align features $F_t(I_t), F_{t-1}(I_{t-1})$, we need to first transform geometric information of $F_{t-1}(I_{t-1})$ to the current frame.
To process $k$ frames simultaneously, we apply a geometric transformation on past features, \ie{} learning $\hat{F}_{t}(I_{t-1})$.
After alignment, we can simply merge two frames
\begin{equation}
\label{equ:2frame_concat}
    \hat{F}_t(I_t,I_{t-1}) = \left[F_t(I_t) \oplus \hat{F}_{t}(I_{t-1})\right].
\end{equation}

The problem is then transformed to how to realize the geometric transformation.
We rephrase the transformation with a perspective of Bayesian estimation.
We inferred that the transformed past features can be described by a conditional probability-based transformation.
\begin{equation}
\label{equ:feature_ego_trans}
    \hat{F}_{t}(I_{t-1}) = \sum_{R_t\in \Omega} p(R_t|I_{t-1}) F_{t-1}(I_{t-1}),
\end{equation}
We expand $p(R_t|I_{t-1})$ with empirical Bayesian
\begin{align}
    \notag p(R_t&|I_{t-1})\\
    \notag&=\int_\Omega p(R_t|R_{t-1}, I_{t-1})p(R_{t-1}|I_{t-1}) \text{ d}R_{t-1}\\
    \label{equ:prob_cascade}
    &=\sum_{R_{t-1}\in\Omega}p(R_t|R_{t-1}, I_{t-1})p(R_{t-1}|I_{t-1}).
\end{align}
From the above equation, we have the following observation: 
\textbf{(i)} $p(R_{t-1}|I_{t-1})$ is the product of the heatmap head of the single-frame detector according to Equation~\ref{equ:single_det}.
\textbf{(ii)} Summation overall space $\Omega$ is memory consuming.
Therefore, a small subspace that can also approximate the equation will be preferred.


\section{Method}
\label{sec:method}
Based on these observations, we propose our multi-frame detector \model{}.
Section~\ref{sec:method_significant_sampling} illustrates the significant sampling approach for the subspace $\Omega_s$, which produce and store a series of sparse BEV features with rich information.
Section~\ref{sec:method_geo_trans_learn} shows the geometric transformation learning that aggregates features from different times.
Finally, we illustrate the overall multi-frame detector in Section~\ref{sec:method_multi_frame_det}.

\subsection{Significant sampling}
\label{sec:method_significant_sampling}
The output of the heatmap head in our single-frame detector, denoted by $p(R_{t-1}|I_{t-1})$, is a sparse distribution. This is because the heatmap only indicates a few locations where objects may be present. Thus, a subspace $\Omega_s\subset \Omega$ with a significant distribution can be identified, where $|\Omega_s|<<|\Omega|$. To save computation resources while approximating the original target, we propose a significant sampling process consisting of three steps: coarse sampling, refined sampling, and relaxation.

\noindent\textbf{Coarse Sampling. }
We use a threshold $\alpha$ to select $R_{t-1}$ which $p(R_{t-1}|I_{t-1}) < \alpha$ in Equation~\ref{equ:prob_cascade}, where $\alpha$ is a small real number called {\em significant threshold}.

\paragraph{Refined Sampling. }
Based on the coarse sampling, we have sampled all the points of interest but many of them actually belong to the same Gaussian peak.
To mitigate this issue, we propose a refinement stage based on Non-maximum Suppression (NMS). Specifically, we leverage the location and scale information obtained from the Regression heads of the single frame detection to filter out overlapping boxes. By applying NMS, we ensure that only the most promising and representative boxes are retained while discarding redundant ones."

\noindent\textbf{Relaxation. }
After refined sampling, we find that simply using the feature $F_{t-1}(I_{t-1})$ exacted at the sampled points loses too much information.
Therefore, we make a neighborhood relaxation $\mathcal{B}(r_{t-1})$ around the sample points $r_{t-1}$.
We try three relaxation methods -- rectangle window, circular window, and square window.
\textit{Rectangle window} takes points inside the box predicted by Regression heads of single frame detector.
\textit{Circular window} takes points from the circle with radius $\rho$ around $r_{t-1}$ .
\textit{Square window} takes points inside the $H_s\times W_s$ square centered at $r_{t-1}$.
our significant sampling sample a minority of locations $\Omega_s\subset \Omega$.
Noted that the scale of $\Omega_s$ ($10^1$) is greatly smaller than that of $\Omega$ ($10^4$).
Now, we can simplify the Equation~\ref{equ:prob_cascade} by sampling only locations with significant influence, \ie
\begin{align}
    \notag &p(R_t|I_{t-1})\\
    \label{equ:sig_prob_cascade}&\approx\sum_{R_{t-1}\in\Omega_s}p(R_t|R_{t-1}, I_{t-1})p(R_{t-1}|I_{t-1}).
\end{align}
Consequentially, combining Equation~\ref{equ:feature_ego_trans} and \ref{equ:sig_prob_cascade}, we get
\begin{align}
    \notag &\hat{F}_{t}(I_{t-1})\\
    \notag&\approx \sum_{R_t\in \Omega} \left[\sum_{R_{t-1}\in\Omega_s}p(R_t|R_{t-1}, I_{t-1})p(R_{t-1}|I_{t-1})\right] F_{t-1}(I_{t-1})\\
    \label{equ:final_multi}
    &=\sum_{R_{t}\in\Omega}p(R_t|R_{t-1}, I_{t-1})\left[\sum_{R_{t-1}\in\Omega_s}p(R_{t-1}|I_{t-1}) F_{t-1}(I_{t-1})\right].
\end{align}
Therefore, only Feature $F_{t-1}(I_{t-1})$ sampled in $\Omega_s$ need to be stored for next frame inference which shows our memory efficiency.

\subsection{Geometric transformation learning}
\label{sec:method_geo_trans_learn}
Geometric transformation can be divided into two guises: ego-car transformation and object-centric transformation.
For ego-car transformation, we directly transform the feature map of the last frame to the current frame BEV coordinate so that $p(R_{t-1}|I_{t-1})$ will be directly predicted under the current BEV coordinate.
This transformation can be carried out at no additional cost. On the other hand, the object-centric transformation is more complex and requires explicit consideration in our method.
We introduce our object-centric transformation learning in two parts: 
theory and implementation.

\paragraph{Theory}
The object-centric transformation is based on the velocities of objects.
To acquire ground-truth velocities of each sample from $\Omega_s$ of past frames, we first assign each sample with the specific ground-truth object.
The velocity of the successfully assigned sample is equal to the velocity of the assigned object while the velocity of the unassigned sample is equal to 0.
We denote the ground-truth velocity for $r_{t-1}$ as $v^{(r_{t-1})}_{t-1}\in \mathbb{R}^2$, then the ground-truth distribution for $p$ can be derived as
\begin{equation}
\label{equ:geo_trans_gt_gl}
    p(\hat{R}_t|r_{t-1}, I_{t-1})=
    \begin{cases}
        1, & \hat{R}_t = r_{t-1} + v^{(r_{t-1})}_{t-1}\cdot \tau \\
        0, & \text{otherwise}.
    \end{cases}
\end{equation}
where $\tau$ is the time gap between frames.
\begin{table*}[t]
	\begin{center}
	\resizebox{\textwidth}{!}{
		\begin{tabular}{c|c|c|cc|cccccccccc} 
			\hline
			Method &Backbone &Information &\textbf{mAP} &\textbf{NDS} & Car & Truck & C.V. & Bus & Trailer & Barrier & Motor. & Bicycle & Ped. & T.C. \\
			\hline\hline
			PointPillars~\cite{lang2019pointpillars} &Pillar &L &30.5 &45.3  & 68.4 & 23.0 & 4.1  & 28.2 & 23.4 & 38.9 & 27.4 & 1.1  & 59.7 & 30.8 \\
			WYSIWYG~\cite{hu2020you} &Pillar &L  &35.0 & 41.9  & 79.1 & 30.4 & 7.1 & 46.6 & 40.1 & 34.7 & 18.2 & 0.1 & 65.0 & 28.8 \\ 
 			3DVID~\cite{yin2020lidar} &Pillar &L+T &45.4 &53.1  & 79.7 & 33.6 &18.1&47.1&43.0&48.8&40.7&7.9&76.5&58.8 \\
 			TCTR~\cite{yuan2020tctr} &Pillar &L+T &50.5 &-  &83.2 &51.5 &15.6 &\textcolor{blue}{\textbf{63.7}} &33.0 &53.8 &54.0 &22.6 &74.9 &52.5 \\
 			TransPillars~\cite{luo2022transpillars} &Pillar &L+T &52.3 &-  &84.0 &\textcolor{blue}{\textbf{52.4}} &18.9 &62.0 &34.3 &55.1 &55.2 &27.6 &77.9 &55.4 \\
            \hline
            \textbf{\model (Ours)} &Pillar &L+T &\textcolor{blue}{\textbf{57.1}} &\textcolor{blue}{\textbf{65.5}} &\textcolor{blue}{\textbf{84.6}} &48.8 &\textcolor{blue}{\textbf{23.9}} &57.5 &\textcolor{blue}{\textbf{53.8}} &\textcolor{blue}{\textbf{63.2}} &\textcolor{blue}{\textbf{58.6}} &\textcolor{blue}{\textbf{30.5}} &\textcolor{blue}{\textbf{80.3}} &\textcolor{blue}{\textbf{69.6}} \\ 
 			\hline
                \hline
 			CBGS~\cite{zhu2019class} &Voxel &L  &52.8 &63.3   &81.1 &48.5 &10.5 &54.9 &42.9 &65.7 &51.5 &22.3 &80.1 &70.9 \\
 			CenterPoint~\cite{yin2020center} &Voxel &L &58.0 &65.5  &84.6 &51.0 &17.5 &\textcolor{red}{\textbf{60.2}} &53.2 &70.9 &53.7 &28.7 &83.4 &76.7 \\
			\hline
			\textbf{\model (Ours)} &Voxel &L+T  &\textcolor{red}{\textbf{62.8}} &\textcolor{red}{\textbf{68.9}} &\textcolor{red}{\textbf{85.9}} &\textcolor{red}{\textbf{53.7}} &\textcolor{red}{\textbf{26.4}} &59.0 &\textcolor{red}{\textbf{54.9}} &\textcolor{red}{\textbf{71.6}} &\textcolor{red}{\textbf{68.9}} &\textcolor{red}{\textbf{42.9}} &\textcolor{red}{\textbf{85.6}} &\textcolor{red}{\textbf{79.7}} \\ 
			\hline
		\end{tabular}
		}
	\end{center}
	\caption{Performance comparisons on the nuScenes test set.
	We report the overal mAP, NDS, and mAP for each detection category, where ``L'' denote LiDAR modality, and T denotes multi-frame Temporal input.
	 ``-'' represents the unknown information.
	 We highlight the best results with pillar-based backbone in \textcolor{blue}{\textbf{blue}} and with voxel-based backbone in \textcolor{red}{\textbf{red}}. 
	}\label{tab-nus-test}
\end{table*}
However, learning the probability across $\hat{R}_t$ on all the point cloud ranges is tedious and difficult.
To tackle the issue, we make the following local approximation.
We limit the space of $R_t$ inside the neighbor of $r_{t-1}$, \ie\ a $H_w\times W_w$ window $\mathcal{B}(r_{t-1})$.
We define a probability $q(\hat{R}_t|r_{t-1}, I_{t-1})$ to represent the ground-truth geometric transformation probability, where $\hat{R}_t$ is limited inside the window $\mathcal{B}(r_{t-1})$.
Assume that, the choice of $H_w, W_w$ can adequately take $v^{(r_{t-1})}_{t-1}\cdot \tau\in\mathcal{B}(r_{t-1})$ for all $r_{t-1}\in\Omega_s$, we have
\begin{equation}
\label{equ:geo_trans_gt_loc}
    q(\hat{R}_t|r_{t-1}, I_{t-1})=
    \begin{cases}
        1, & \hat{R}_t = v^{(r_{t-1})}_{t-1}\cdot \tau \\
        0, & \text{otherwise}.
    \end{cases}
\end{equation}
Then we define the $q_\phi(R_t|r_{t-1}, I_{t-1})$ to represent the predicted geometric transformation probability, where $R_t$ is also limited inside the window $\mathcal{B}(r_{t-1})$.
During optimization, we minimize the divergence between probability, \ie
\begin{equation}
    \min_\phi \mathcal{D}(q(\hat{R}_t|r_{t-1}, I_{t-1}), q_\phi(R_t|r_{t-1}, I_{t-1}))
\end{equation}
Finally, the global predicted geometric transformation probability can be approximated with local probability by
\begin{align}
\tiny
    p(R_t|r_{t-1}&, I_{t-1})=
    &\begin{cases}
        q(R_t|r_{t-1}, I_{t-1}), & R_t\in \mathcal{B}(r_{t-1}) \\
        0, & \text{otherwise} .
    \end{cases}
\end{align}

\paragraph{Implementation}
We predict geometric transformation probability of position $r_{t-1}\in\Omega_s$ from the feature $F_{t-1}(I_{t-1})$ sampled around $r_{t-1}$.
Given relaxation form Section~\ref{sec:method_significant_sampling}, we can acquire a $ H_s\times W_s\times C$ local feature map.
Then we use a simple three-layer CNN called \textbf{GeoNet} on the local feature map to produce the geometric transformation probability $1\times 1\times (H_w W_w)$.
During the training phase, we use Cross Entropy Loss to optimize the distribution. 
\subsection{Detection Head}
\label{sec:method_multi_frame_det}
Finally, Equation~\ref{equ:2frame_concat} gives the aligned aggregation.
For a longer sequence, we use induction. 
Taking $t+1$ for example, we can replace $F_{t}(I_{t})$ with $\hat{F}_{t}(I_{t-1})$, \ie{}
\begin{equation}
    \hat{F}_{t+1}(I_{t+1},I_{t}) = f_2\left[F_{t+1}(I_{t+1}), \hat{F}_{t}(I_{t-1})\right].
\end{equation}
After the acquirement of the temporal unified feature, the following detection heads -- heatmap head and regression head are just the same as the single-frame 3D detector.

\section{Experiments}
\label{sec:experiments}
We evaluate our method on two widely used large-scale datasets including nuScenes~\cite{caesar2020nuscenes} and Waymo~\cite{sun2020scalability}.
We conduct extensive ablation studies to validate our design choices.
\begin{table}[t]
	\begin{center}
	\resizebox{0.48\textwidth}{!}
{
		\begin{tabular}{c|c|cc|cc}
			\hline
			Method &Frames &\textbf{mAP} &\textbf{NDS} &Latency(ms) &Memory(GB) \\
			\hline
			Centerpoint~\cite{yin2020center} &1  &59.6 &66.8 &\textbf{78.6} &\textbf{5.9} \\
			INT~\cite{xu2022int} &1  &58.5 &65.5  &84.1 &12.1\\
			INT~\cite{xu2022int} &2  &60.9 &66.9  &84.1 &12.1\\
			INT~\cite{xu2022int} &10 &61.8 &67.3  &84.1 &12.1 \\
			\hline
			\textbf{\model (Ours)} &2  &\textbf{63.1} &\textbf{68.4} &107.5 &6.2\\
                \textbf{\model (Ours)} &10  &\textbf{63.7} &\textbf{68.6} &107.5 &6.2\\
			\hline
		\end{tabular}
		}
	\end{center}
	\caption{Performance comparisons on the nuScenes validation set.
	We report the overal mAP, NDS.
 All models are built on the VoxelNet backbone.
	We highlight the best results in \textbf{blod}.
	}\label{tab-nus_val}
\end{table}

\subsection{Experimental Setup}
\noindent \textbf{Dataset.}
The nuScenes~\cite{caesar2020nuscenes} is a large-scale dataset for autonomous driving. 
It contains 1,000 driving sequences, which are divided into 700, 150, and 150 for training, validation, and testing, respectively. 
Each sequence lasts about 20 seconds, with a LiDAR sampling frequency of 20HZ.
nuScenes provides annotations every 0.5 seconds including 10 classes with a long-tail distribution.
The Waymo Open Dataset~\cite{sun2020scalability} is consisting of 798 training scenes and 202 validation scenes. 
Data collection and 3D annotation are both at 10 Hz frequency. Notably, nuScenes annotates 10 classes for 3D detection task, while Waymo provides annotations of 3 classes for its 3D detection benchmark.

\begin{table}[t]
	\begin{center}
	\resizebox{0.48\textwidth}{!}
{
		\begin{tabular}{c|c|cccc|c}
    			\hline
                 Methods &Frames & VEH &PED &CYC &\textbf{mAPH} &Latency(ms)\\
    			\hline
                    CenterPoint ~\cite{yin2020center} &1 &66.2 &62.6 &67.6 &65.5 &71.7\\
                    CenterPoint ~~\cite{yin2020center} &2 &67.3 &67.5 &69.9 &68.2 &90.9\\
                    3D-MAN ~\cite{20213D} &15 &67.1 &59.0 &- &- &-\\
                    RSN ~\cite{sun2021rsn} &3 &69.1 &- &- &- &\textbf{67.5} \\
                    INT ~\cite{2021Towards} &2 &69.4 &69.1 &\textbf{72.6} &70.3 &74.0\\
                    \hline
    			\textbf{SUIT} &2 & \textbf{70.0} & \textbf{70.3}  &72.4 &\textbf{70.9} &79.3\\
    			\hline
    		\end{tabular}
		}
	\end{center}
	\caption{Performance comparisons on the Waymo validation set.
	We report the LEVEL\_2 mAPH(\%) for each category. 
	All models are built on the VoxelNet backbone.
 We highlight the best results in \textbf{blod}.
	}\label{tab-waymo_val}
\end{table}
\noindent \textbf{Experimental Details.}
We follow CenterPoint~\cite{yin2020center} to set our backbone network.
For nuScenes data, we set the range of the point cloud space as $[-54.0m, 54.0m]$ for the $X$ and $Y$ axes, and $[-5m, 3m]$ for the $Z$ axis.
The size of each voxel is set to $(0.075m, 0.075m, 0.2m)$.
For Waymo data, we set the range of the point cloud space as $[-75.2m, 75.2m]$ for the $X$ and $Y$ axes, and $[-2m, 4m]$ for the $Z$ axis.
The size of each voxel is set to $(0.1m, 0.1m, 0.15m)$.
We trained our model using Adam optimizer with one cycle learning rate policy, weight decay 0.01, and momentum 0.85 to 0.95 on 8 Telsa V100 GPUs.
The total batch size of our model is 32. 
We use the official CenterPoint~\cite{yin2020center} checkpoints as our pretrained model and fine-tune 6 epochs.
During training, we use random flipping, global scaling, global rotation, and global translation as our data augmentation.

\begin{figure*}[t]
	\begin{center}
		\includegraphics[width=1.0\linewidth]{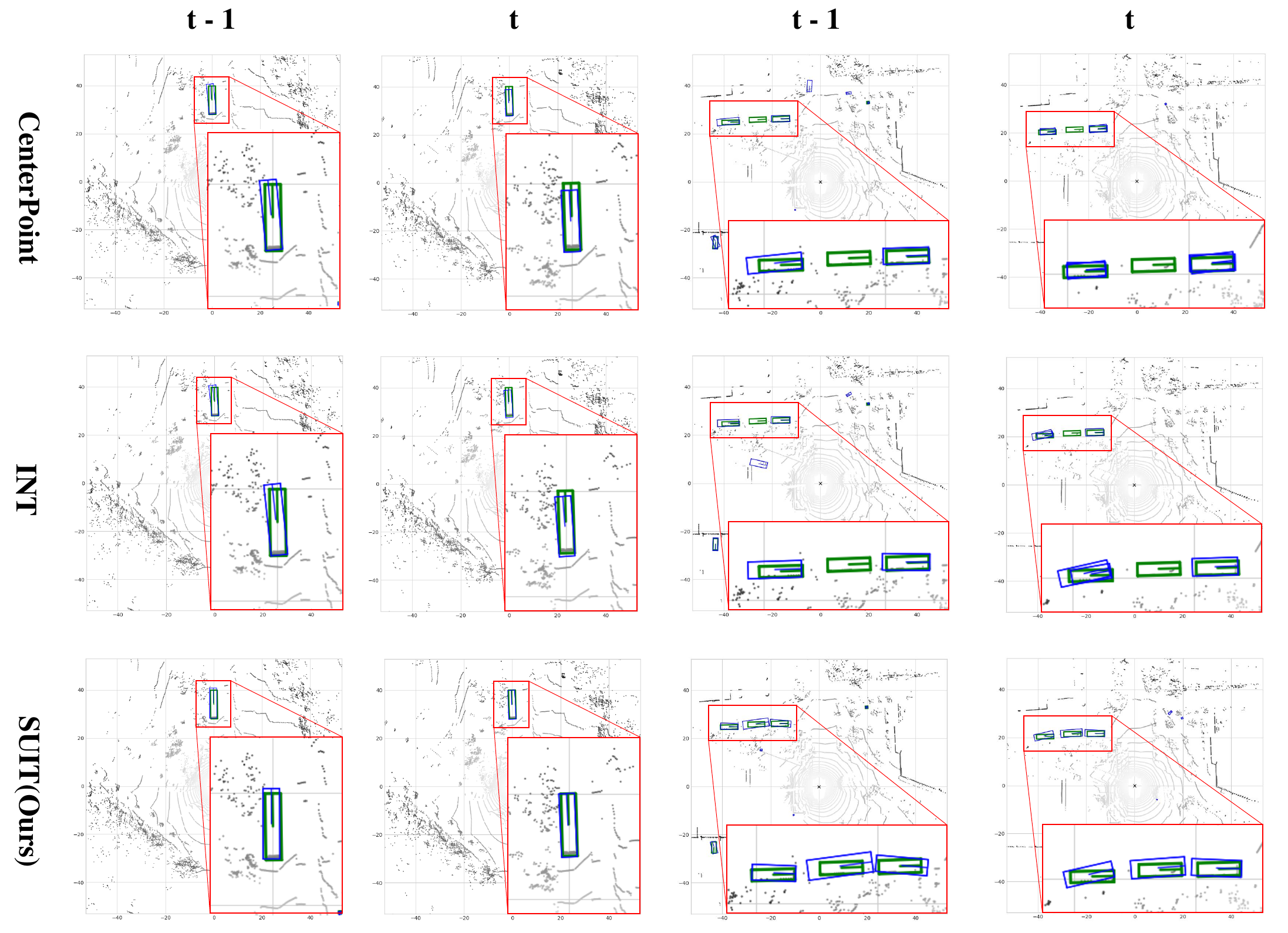}
	\end{center}
	\caption{Qualitative comparison of temporal 3D detector between INT~\cite{xu2022int} ({\bf \em middle}) and ours ({\bf \em bottom}) on nuScenes dataset.
	We also provide the single-frame baseline CenterPoint~\cite{yin2020center}({\bf \em top}) to show its detection result.
	Boxes in \textcolor{blue}{blue} are the predicted results while the \textcolor{green}{green} boxes are the ground truth.
	}
	\label{fig-vis-nus}
\end{figure*}
\paragraph{\bf Metrics.}
We use mean Average Precision (mAP)~\cite{everingham2010pascal} and NuScenes Detection Score (NDS)~\cite{caesar2020nuscenes} to evaluate our performance.
The mAP stands for the average of precision on 0.5m, 1m, 2m, 4m four different distance thresholds. 
NDS is a particular metric on nuScenes~\cite{caesar2020nuscenes} dataset. 
It is the average weighted summation of mAP and True Positive (TP) metrics.
TP metrics average five individual metrics: translation (ATE), velocity (AVE), scale (ASE), orientation (AOE), and attribute (AAE) errors.
The weighted average is calculated by: $\text{NDS} = \frac{1}{10}\left[ 5\text{mAP} + \sum_{\text{mTP}\in\mathbb{TP}} (1-\min(1, \text{mTP}))  \right]$.
We follow the official evaluation metrics mAP and mAPH (mAP weighted by heading) on Waymo Open Dataset. 
The metrics are further split into two difficulty levels according to the point numbers in GT boxes: LEVEL\_1 ($>$5) and LEVEL\_2 ($>=$1).

\subsection{Main Results}
\paragraph{nuScenes Results. }
Table~\ref{tab-nus-test} shows our comparison with state-of-the-art single-frame and multi-frame methods on nuScenes~\cite{caesar2020nuscenes} \texttt{test} split. Since nuScenes provides 10 samples between annotated frames, we follow the common technique in INT~\cite{xu2022int} which aggregates the point cloud of those interpolated samples to represent one frame.
We first compare our multi-frame improvement with the single-frame baseline Centerpoint~\cite{yin2020center}.
According to the $8^{th}$ and $9^{th}$ row, our multi-frame mechanism bring $4.8$ point improvement on mAP and $3.4$ improvement on NDS.
To compare with other state-of-the-art methods, we mainly select  pillar-based methods for the lack of voxel-based multi-frame results on nuScenes~\cite{caesar2020nuscenes} dataset.
Comparing the $5^{th}$ and $6^{th}$ row, we surpass the former multi-frame state-of-the-art TransPillars~\cite{luo2022transpillars} by $4.8$ point on mAP.
3DVID~\cite{yin2020lidar}, TCTR~\cite{yuan2020tctr} and  TransPillars~\cite{luo2022transpillars} utilize implicit GRU or Transformer on dense features to learn a geometric transformation.
Our higher performance proves the importance of explicit supervision of geometric transformation.

Table~\ref{tab-nus_val} shows our comparison with state-of-the-art multi-frame methods on nuScenes~\cite{caesar2020nuscenes} \texttt{validation} split. 
On the 2-frame paradigm, we outperform the former state-of-the-art INT~\cite{xu2022int} with $2.2$ point on mAP, $1.5$ point on NDS under the voxel-based backbone.
On a longer sequence, our 10-frame detector surpasses 10-frame detector of INT~\cite{xu2022int} by $2.8$ point on mAP under the voxel-based backbone.
Our \model{} achieves the best mAP and the best memory cost among multi-frame models. As the frame length increases, \model{} still runs with low latency and memory cost, benefiting from the sparse information.
INT~\cite{xu2022int} use a loose uniform distribution assumption when handling object-centric geometric transformation.
The loose assumption, according to the comparison, harms temporal information learning.
Therefore, our explicit geometric transformation learning drastically surpasses the INT~\cite{xu2022int}.

\paragraph{Waymo Resutls.}
We also make comparisons on the Waymo dataset as in Table~\ref{tab-waymo_val}. Note that nuScenes and Waymo differ a lot in terms of sensor configurations and evaluation classes, the detection model is hard to be reused among datasets, which is a consensus in previous work. Thus we train a detector on the Waymo dataset with the same model architecture on the nuScenes dataset.
\begin{table}[h]

\footnotesize
	\begin{center}
  {
    		\begin{tabular}{c|cc|cc}
    			\toprule
    		    Top K & $\alpha$ & Refined &\textbf{NDS} &\textbf{mAP} \\
    		    \hline
    		   500  &-      &-              &67.7 &62.6 \\
    		   200  &-      &-              &68.1 &62.6 \\
    		   200  &0.1    &-              &68.2 &62.7 \\
    		   200  &0.2    &-              &68.2 &62.7 \\
    		   200  &0.1    &\checkmark     &\textbf{68.4} &\textbf{63.1}  \\
     \bottomrule
    		\end{tabular}
}
	\end{center}
	\caption{Ablations on the sampling schemes. 
  We highlight the best results in \textbf{blod}.
	} \label{tab-ablation-sample1}

\end{table}
\begin{table}[t]
\footnotesize
	\begin{center}
  {
    		\begin{tabular}{cc|cc}
    			\toprule
    		    Windows & Relaxation &\textbf{NDS} &\textbf{mAP} \\
    		    \hline
    		   Rectangle    & -                     &65.9 &61.3 \\
    		   Circular      &$\rho$=2              &68.0 &62.9 \\
    		   Square       &$H_s=W_s$=3            &68.4 &\textbf{63.1} \\
    		   Square       &$H_s=W_s$=5            &67.5  &62.9 \\
    		   Square       &$H_s=W_s$=7            &\textbf{68.5}  &62.9 \\

     \bottomrule
    		\end{tabular}
}
	\end{center}
	\caption{Ablations on the sampling relaxation. 
  We highlight the best results in \textbf{blod}.
	} \label{tab-ablation-sample2}
\end{table}
The results in Table~\ref{tab-waymo_val} show that our model consistently outperforms previous methods and boosts the detector by on the three challenging classes for the Waymo evaluation benchmark. 
Experimental results demonstrate that our method generalizes and scales well on both datasets with state-of-the-art performance.

\subsection{Qualitative Results. }
\paragraph{Multi-frame detection results.} 
We present visualizations of multi-frame detection results in Figure~\ref{fig-vis-nus}.
We compare single-frame detection baseline Centerpoint~\cite{yin2020center} and the previous state-of-the-art multi-frame detector INT~\cite{xu2022int}.
Compared with the single-frame detector, we can refine the current frame detection results by the last frame.
Box circled in red notation is refined based on the last frames by our method, but the single-frame cannot build the temporal refinement.
Compared with the previous, we predict a more accurate result under a lower cost on memory.

\subsection{Ablation Studies}
\label{sec:abl_sampling}
\paragraph{Analysis on sample stages. }
Table~\ref{tab-ablation-sample1} shows the investigation of two stages of significant sampling -- coarse sampling and refined sampling.
First, we do not use coarse sampling and refined sampling but only leave samples with top-K probability. 
Comparing the $1^{st}$ and $2^{nd}$ rows, we observe that leaving the top 500 samples performs worse than the top 200 although more samples reserve more information during sampling.
It shows that introducing more noise object-centric samples disturb the following object-centric geometric transformation learning.
On the other hand, when we only use coarse sampling, we tune the significant threshold $\alpha$. 
Comparing the $3^{rd}$ and $4^{th}$ rows, we observe that a larger threshold, filtering more samples, however, leads to poorer performance.
It illustrates that, on the other hand, too less samples can omit too many significant samples thus weakening our model.
Finally, we compare the importance of refined sampling between $3^{rd}$ and $5^{th}$ rows.
With refined NMS operation, we can whittle down noise samples thus helping the geometric transformation learning.

\begin{table}[t]
\footnotesize
	\begin{center}
		\resizebox{0.48\textwidth}{!}
		{
    		\begin{tabular}{ccc|cc}
\toprule
$F_{t-j}(I_{t-j})$ & $p_\theta(R_{t-j}|I_{t-j})$ & $p_\phi(\tilde{R}_t|R_{t-j},I_{t-j})$  &\textbf{NDS} &\textbf{mAP}\\
\hline
\XSolidBrush       & \XSolidBrush & \XSolidBrush &66.5  &60.9  \\
\checkmark & \XSolidBrush & \XSolidBrush             &68.1  &62.6  \\
\checkmark & \checkmark & \XSolidBrush   & 68.1 &62.6\\
\checkmark & \checkmark& \checkmark& \textbf{68.4} & \textbf{63.1}\\

     \bottomrule
    		\end{tabular}
}
	\end{center}
	\caption{Ablations on the geographic transformation components.
  We highlight the best results in \textbf{blod}.
	} \label{tab-ablation-geo1}
\end{table}

\paragraph{Analysis on relaxation. }
Table~\ref{tab-ablation-sample2} studies three relaxation methods.
We have the following observations. 
(i) Rectangle window in the $1^{st}$ achieves the worst performance. 
The potential reason is the error brought by Regression heads since it uses the predicted box as a neighbor.
(ii) Circular windows on $2^{nd}$ and $3^{rd}$ rows achieve better than rectangle windows and are similar to square windows.
The circular windows have an irregular boundary, thus less efficient than square windows
(iii) Square windows perform the best on the mAP metric.
Also, the regular boundary is more efficient when accessing memory-storing features.
The different window sizes can achieve similar performance but considering the efficiency of small window sizes, we finally choose square window relaxation with the window size equal to $3$. 

\begin{table}[t]
\footnotesize
\begin{center}
{
    \begin{tabular}{c|c|cc}
        \hline
             &Components & \textbf{NDS} &\textbf{mAP} \\
        \hline
            0 &Baseline  &66.8 &59.6 \\
            \hline
            \multirow{2}{*}{1} &+dense BEV &67.9 &60.8\\
            &+GeoNet &68.2 &62.9 \\
            \hline
            \multirow{2}{*}{2} &+sparse BEV &68.1 &62.6 \\
            &+speed &67.8 &62.2 \\
            \hline
            \multirow{2}{*}{3} &+sparse BEV &68.1 &62.6 \\
            &+GeoNet &\textbf{68.4} &\textbf{63.1}\\
        \hline
    \end{tabular}
}
\end{center}
\caption{Ablation on the GeoNet.
 We highlight the best results in \textbf{blod}.}
\label{tab-ablation-geo2}
\end{table}
\begin{table}[t]
\footnotesize
	\begin{center}
  {
    		\begin{tabular}{c|c|cc}
    			\toprule
    		     Sampling relaxation & Local window &\textbf{NDS} &\textbf{mAP}\\ 
    		    \hline
    		   $H_s=W_s=3$ &$H_w=W_w=3$     &68.2 &62.59 \\
    		   $H_s=W_s=3$ &$H_w=W_w=7$     &\textbf{68.4} &\textbf{63.1} \\
    		   $H_s=W_s=3$ &$H_w=W_w=15$    &68.4 &63.0 \\
    		   $H_s=W_s=7$ &$H_w=W_w=7$     &68.3 &62.8 \\ 
    		   $H_s=W_s=7$ &$H_w=W_w=15$    &68.3 &62.9\\ 
     \bottomrule
    		\end{tabular}
}
	\end{center}
	\caption{Ablations on the geographic learning local windows. Also, we consider the influence of sampling relaxation.
  We highlight the best results in \textbf{blod}.
	} \label{tab-ablation-learning2}
\end{table}

\paragraph{Analysis on components. }
Table~\ref{tab-ablation-geo1} shows the ablation studies of components in Equation~\ref{equ:final_multi}.
Based on the $1^{st}$ row, when feature $F_{t-j}(I_{t-j})$ attending, we find the importance of saving features from the last frames.
Based on the $2^{nd}$ row, when the prior $p(R_{t-j}|I_{t-j})$ attending, we observe a tiny promotion.
It shows that a uniform prior distribution is strong enough.
Comparing the $3^{rd}$ and $4^{th}$ row, a huge difference is made by geometric transformation.
By considering $p_\phi(\tilde{R}_t|R_{t-j},I_{t-j})$, we find a $0.6\%$ on mAP, which proves the significance of geometric transformation.

\paragraph{Analysis on GeoNet}
Table~\ref{tab-ablation-geo2} presents the ablation study results that demonstrate the effectiveness of the proposed GeoNet. In the absence of the proposed significant sampling process, our GeoNet aligns all possible targets on the entire dense BEV feature map, resulting in a modest 2.1 point increase in mAP. A comparison of the last two rows indicates that directly utilizing predicted velocities to align objects fails to enhance the detector's performance. This observation suggests that inaccurate velocity predictions can adversely affect the performance of multi-frame detectors.

Table~\ref{tab-ablation-learning2} shows additional analyses on GeoNet.
We first study the influence of local window size.
Comparing the first three rows, we observe that both too-large and too-small local windows downgrade the geometric learning performance.
The small local window can miss the geometric transformation of long-distance (\eg{} fast vehicles) but is easy to train.
Contrariwise, the large local window can cover larger geometric transformations but is harder to train.
$H_w=W_w=7$ achieves an equipoise between large and small local windows.
The study of sampling relaxation receives a similar observation as that in Section~\ref{sec:abl_sampling}.
There is no significant influence on relaxation size, but smaller sizes are more efficient.

\section{Conclusion}
\label{sec:conclusion}

In this paper, we present a single-stage 3D detection framework \model{}, which aims to exploit the potential of temporal information.
Particularly, we extract the slight sparse features from redundant temporal information, which are then fused across frames to boost detection accuracy under low memory cost. 
Furthermore, we introduce an explicit geometric transformation that learns the relationships throughout the trajectory of an object to enhance the alignment. 
With the proposed end-to-end detector, \model{} achieves significant improvements over strong baselines on the challenging nuScenes and Waymo benchmark datasets.
It is noteworthy that our model is a general multi-frame approach that can be extended to other related tasks such as segmentation and tracking.

\bibliographystyle{IEEEtran}
\bibliography{IEEEfull}

\end{document}